\documentclass{article}
\usepackage{spconf,amsmath,graphicx}

\title{Guided Filter based Edge-preserving Image Non-blind Deconvolution}
\twoauthors
  {Hang Yang, Ming Zhu}
	{Changchun Institute of Optics, Fine Mechanics and Physics\\
	Chinese Academy of Science\\
	China, Changchun 130033\\ yanghang09@mails.jlu.edu.cn\\ mingzhuca@163.com}
{Zhongbo Zhang, Heyan Huang}
	{Department of Mathematics\\
	Jilin University\\
	China, Changchun 130012\\ zhongbozhang@jlu.edu.cn \\ huanghy10@mails.jlu.edu.cn}
\begin{document}
%\ninept
%
\maketitle
\begin{abstract}
In this work, we propose a new approach for efficient edge-preserving image
deconvolution. Our algorithm is based on a novel type of explicit image filter - guided filter. The guided filter can be used as an edge-preserving smoothing operator like the popular bilateral filter, but has better behaviors near edges.
We propose an efficient iterative algorithm with the decouple of deblurring and denoising steps in the restoration process.
In deblurring step, we proposed two cost function which could be computed with fast Fourier transform efficiently. The solution of the first one is used as the guidance image, and another solution will be filtered in next step. In the denoising step, the guided filter is used with the two obtained images for efficient edge-preserving filtering. Furthermore, we derive a simple and effective method to automatically adjust the regularization parameter at each iteration.  We compare our deconvolution algorithm with many competitive deconvolution techniques in terms of ISNR and visual quality.
\end{abstract}
\section{Introduction}
\label{sec:intro}

Image deconvolution is a classical inverse problem existing in a wide variety of image processing fields, including physical,
optical, medical, and astronomical applications.

The degradation procedure is often modeled as the result of a convolution with a low-pass filter
\begin{equation}\label{1.1}
    y= \mathcal{H}u_{orig}+ \gamma  =h \ast u_{orig}+\gamma
\end{equation}
where $u_{orig}$ and $y$ are the original image and the observed image, respectively. $\gamma$ is generally assumed to be independent
and identically distributed (i.i.d.) zero-mean additive white Gaussian noise (AWGN) with variance $\sigma^{2}$. $"\ast"$ denotes convolution, and $h$ denotes the point spread function (PSF) of a linear time-invariant (LTI) system $\mathcal{H}$.

To find a unique and stable solution, a number of deconvolution algorithms have been proposed. In these methods, the Wiener filter and the constrained least squares algorithm, can solve this problem in the frequency domain in a fast speed. In \cite{R.Neelamani}, Neelamani $et al.$ proposed an efficient, hybrid Fourier-wavelet regularized deconvolution (ForWaRD) algorithm. Transformations such as curvelets \cite{J.Starck}, shearlets \cite{VishalM.Patel} and wave atoms \cite{Hang} are popular for image representation and are often
used for image restoration. Another popular deconvolution method is based on total variation. Variations of this method have also been proposed in \cite{Y.Wang}\cite{OlegV.Michailovich}. These methods are well known for its edge-preserving property, and can generally achieve state-of-the-art results. In particular, the SV-GSM \cite{J.A.Guerrero-Colon} and the BM3D (Block Matching 3D) \cite{Kostadin Dabove} are among the current best image deconvolution methods. There are many useful algorithms and additional techniques in references \cite{R.Rubinstein}\cite{J.Portilla}.

In this work, we adopt a different approach to the problem of image
restoration by exploiting guided filter \cite{He} to regularize the inverse
problem. Derived from a local linear model, guided filter generates the filtering output by
considering the content of a guidance image. We first integrate this filter into a iterative deconvolution method.
The iterative process consists of two parts: debluring and denoising. The output of the debluring process are one noisier estimated image and a less noisy one. The former will be filtered  and the latter will work as the guidance image respectively in denoising step. During the denoising process, the guided filter will be applied to the output of last step to reduce noise and refine the result of last step.
Furthermore, regularization parameters play the important role in our method. We apply the discrepancy principle to automatically determine regularization parameters in each iteration.
We demonstrate with experimental results that this algorithm provides competitive and even better figures of merit compared with state-of-the-art methods.

\section{Guided Image Filtering}

Guided filter was defined in \cite{He}\cite{He1}.  Currently it is one of the fastest edge-preserving filters.
Now, we introduce guided filter, which involves a guidance image $u_{I}$, an filtering input image $u_{p}$, and an output image $u$.
Both $u_{I}$ and $u_{p}$ are given beforehand according to the application, and they can be identical.

The key assumption of the guided filter is a local linear model between the
guidance  $u_{I}$ and the filtering output $u$. We assume that $u$ is a linear transform of  $u_{I}$ in a window $\omega_{k}$ centered
at the pixel $k$ (the size of $\omega_{k}$ is $w\times w$.) :
\begin{equation}\label{2.1}
    u(i) = a_{k} u_{I}(i) + b_{k}
\end{equation}
where $(a_{k}, b_{k})$ are some linear coefficients assumed to be constant in $\omega_{k}$.  They can be computed as:
\begin{eqnarray}
  a_{k} &=& \frac{\frac{1}{w^{2}}\sum_{i \in \omega_{k}}u_{I}(i)u_{p}(i)-\mu_{k}\bar{p}_{k}}{\sigma^{2}_{k}+\varepsilon} \\
  b_{k} &=& \bar{p}_{k} - a_{k}\mu_{k}
\end{eqnarray}

%To determine the linear coefficients $(a_{k}, b_{k})$, we need constraints from the filtering input $p$. Specifically, we minimize the following cost function in the window $\omega_{k}$
%\begin{equation}\label{2.2}
%    E(a_{k}, b_{k}) = \sum_{i \in \omega_{k}} ((a_{k}I_{i}+b_{k}-p_{i})^{2} + \varepsilon a^{2}_{k})
%\end{equation}
%Here $\varepsilon$ is a regularization parameter penalizing large $a_{k}$.
%The solution of Eq.(\ref{2.2}) can be written as:
Here, $\mu_{k}$ and $\sigma_{k}$ are the mean and variance of $u_{I}$ in $\omega_{k}$, and $\bar{p}_{k}$ is the mean of $u_{p}$ in $\omega_{k}$. %Having obtained the linear coefficients $(a_{k}, b_{k})$, we can compute the filtering output $q_{i}$ by Eq.(\ref{2.1}).

However, a pixel $i$ is involved in all the overlapping windows $\omega_{k}$ that covers $i$, so the filtering output $u(i)$ can be computed by:%so the value of $q_{i}$ in Eq.(\ref{2.1})
%is not identical when it is computed in different windows. A simple strategy is to average all the possible values of $q_{i}$.
%So, we compute the filtering output by:
\begin{equation}\label{2.3}
    u(i) = \bar{a}_{i}u_{I}(i) + \bar{b}_{i}
\end{equation}
where $\bar{a}_{i} = \frac{1}{w^{2}}\sum_{k \in \omega_{k}}a_{k} $ and $\bar{b}_{i} = \frac{1}{w^{2}}\sum_{k \in \omega_{k}}b_{k}$ are the average coefficients of all windows overlapping $i$. More details and analysis can be found in \cite{He1}.

We denote the Eq.(\ref{2.3}) as $u = \textbf{guidfilter}(u_{I},u_{p},w,\varepsilon)$.

\section{Guided Image Deconvolution}

\subsection{Proposed Deconvolution Algorithm}

Our algorithm is based on the decouple of deblurring and denoising
steps in the restoration process.

In the \textbf{deblurring} step, we proposed two cost functions:
\begin{eqnarray}
\hspace*{-0.5cm} u_{I} &=& \arg \min_{u} \{ \lambda \parallel \nabla u- \nabla u_{E} \parallel^{2}_{2} + \parallel h \ast u -y  \parallel^{2}_{2}\}\\
\hspace*{-0.5cm}  u_{p} &=& \arg \min_{u}\{ \lambda \parallel u- u_{E} \parallel^{2}_{2} + \parallel h \ast u -y  \parallel^{2}_{2}\}
\end{eqnarray}
where $u_{E}$ is a pre-estimated image, and $\lambda > 0$ is the regularization parameter.

Alternatively, we diagonalized derivative operators after Fast Fourier Transform (FFT) for speedup. These yield solutions in the Fourier domain
\begin{eqnarray}
  \mathcal{F}(u_{I}) &=& \frac{\mathcal{F}(h)^{*}\cdot \mathcal{F}(y)+\lambda \mid\mathcal{F}(\nabla)\mid^{2} \cdot \mathcal{F}(u_{E})}{\mid \mathcal{F}(h) \mid^{2}+\lambda\mid \mathcal{F}(\nabla)\mid^{2}}  \\
  \mathcal{F}(u_{p}) &=& \frac{\mathcal{F}(h)^{*}\cdot \mathcal{F}(y)+\lambda \mathcal{F}(u_{E})}{\mid \mathcal{F}(h) \mid^{2}+\lambda}
\end{eqnarray}
where $\mathcal{F}$ is the FFT operator and $\mathcal{F}(\cdot)^{*}$ denotes the complex conjugate. $| \mathcal{F}(\nabla)|^{2} = |\mathcal{F}(\partial_{x})|^{2} +|\mathcal{F}(\partial_{y})|^{2}$ denotes the Fourier transform of $\nabla$ operator.
The plus, multiplication, and division are all component-wise operators.

To suppress the amplified noise and artifacts introduced by Eq.(9), in the \textbf{denoising} step, we applied the guided filter to smooth the estimated image $u_{p}$, and $u_{I}$ is
used as the guidance image. After the Fourier shrinkage steps [see Eq.(8) and (9)], the image $u_{p}$ contains the more leaked noise and more details than $u_{I}$.
So we use $u_{I}$ as the guidance image and $u_{p}$ as the filtering input image to recover some details and reduce the leaked noise.

The guided filter output is locally a linear transform of the guidance image.
This filter has the edge-preserving smoothing property like the bilateral filter, but does not suffer from the gradient reversal artifacts.
So we integrated this filter into the deconvolution problem. This leads to a powerful algorithm that produces high quality results.

Moreover, the guided filter has a fast and non-approximate linear-time algorithm, whose computational complexity is independent of the filtering kernel size. It has an $O(N^{2})$ time (in the number of pixels $N^{2}$) exact algorithm for both gray-scale and color images.

We summarize the proposed algorithm as follows :

---------------------------------------------------------

$Step~0$: Set $k=0$,  pre-estimated image $u^{k}_{E}=0$, choose guided filter parameters $w$ and $\varepsilon$.

$Step~1$: Use $u^{k}_{E}$ to obtain the filtering input image $u^{k}_{p}$ and the guidance image $u^{k}_{I}$ with the Eq.(9) and Eq.(10), respectively.

$Step~2$: Apply guided filter to $u^{k}_{p}$ with the guidance image $u^{k}_{I}$, and obtain a filtered output $u^{k+1} = \textbf{guidfilter}(u^{k}_{I},u^{k}_{p},w,\varepsilon)$.

$Step~3$: Set $u^{k}_{E}=u^{k+1}$, and $k=k+1$.

~~~ $ back~to~Step~1 $

-------------------------------------------------------

\subsection{Choose Regularization Parameter}

Note that the Fourier-based regularized inverse operator in Eq.(8) and (9), and the
deblurred images depend greatly on the degree of regularization which is determined by the regularization parameter $\lambda$. Now, we describe a simple but effective method to compute the parameters automatically.

Based on Morozov¡¯s discrepancy principle \cite{S. Anzengruber}, which selects $\lambda$ by matching the norm of the residual to some upper bound, a good regularized solution $u$ should lie in the set $ \{u; \parallel h \ast u -y  \parallel^{2}_{2}\leq c^{2}\}$, where $c$ is a constant that depends on the noise level \cite{M.Ng}, we use the set $\mathcal{K}=\{u;\parallel h \ast u -y  \parallel^{2}_{2}\leq \rho N^{2}\sigma^{2}, 0 < \rho \leq 1\}$ in this work. By Parseval's theorem and Eq.(10)
\begin{equation}\label{3.7}
\begin{array}{lll}
    \parallel h \ast u_{p} -y  \parallel^{2}_{2}&=&\parallel \frac{\lambda(\mathcal{F}(h)\cdot \mathcal{F}(u_{E})-\mathcal{F}(y))}{|\mathcal{F}(h)|^{2}+\lambda} \parallel^{2}_{2} \\
    &\leq& \parallel h \ast u_{E} -y  \parallel^{2}_{2}
\end{array}
\end{equation}

If the pre-estimated image $u_{E} \in \mathcal{K}$, we set $u_{I} = u_{p} = u_{E}$, and $\lambda = \infty $; Otherwise, a proper parameter $\lambda$ is chosen by
\begin{equation}\label{3.8}
    \parallel \frac{\lambda(\mathcal{F}(h)\cdot \mathcal{F}(u_{E})-\mathcal{F}(y))}{|\mathcal{F}(h)|^{2}+\lambda} \parallel^{2}_{2} = \rho N^{2}\sigma^{2}
\end{equation}

Notice that the left-hand side is monotonically increasing function in $\lambda$, hence there exist
a unique solution $\lambda$, which can be determined via bisection.

From the Eq.(\ref{3.8}), it is clear that the $\lambda$ increases with the increase of $\rho$. Typically the value of $\rho$ is set to 1 \cite{M.Ng}. But in practice, we find that the large $\lambda \ (\rho = 1)$ often causes a noisy result with
ringing effects, though it can substantially reduce the noise variances. So,
we should choose a smaller $\lambda \ (\rho < 1)$ which would obtain an edge preserving image with less noise.
Then, in the denoising step, our effective approach based on guided filter can be employed.

For a smooth image which contains a little high-frequency information, a large $\rho$ will not produce the strong ringing effects and could
substantially suppress the noise. That is to say, the parameter $\rho$ should increase with the decrease of image variance. According to this property, we compute the $\rho$ as follow:
\begin{equation}\label{3.5}
    \rho=\sqrt{1-\frac{\parallel y-\mu(y) \parallel_{2}^{2}-N^{2}\sigma^{2}}{\parallel h \parallel_{1}^{2}\parallel y \parallel_{2}^{2}}}
\end{equation}
where $\mu(y)$ denotes the mean of $y$.

%Concerning the algorithm complexity we note that the algorithm is fast as it is based on fast guided filter. We start by analyzing the cost of each iteration. The $Step~1$ can be carried out with cost $O(N^{2}\log N)$ using the FFT algorithm; In $Step~2$, the cost of guided filter is $O(N^{2})$. And the cost of computing the parameter $\lambda$ is also low because it is to use FFT algorithm.
%

\section{Experiments Results}
The standard $256\times 256$ images $Cameraman$ and $House$ are tested in our experiments. The experiment settings are presented in Table.\ref{table.1}.  Table.\ref{table.1} describes the different point spread functions (PSF) and different amounts of white Gaussian additive noise.

In the experiments, we work on gray-scale images with intensity values normalized to the range [0,1]. Parameters $\varepsilon $ and $w$ of guided filter (in Eq.(3)) should be set carefully. For this work, we have just tried to evaluate how robust the algorithm behaves for two different images and a wide range of image degradations. For this purpose we have hand-optimized the two parameters for using them with five degradations and two images, obtaining $\varepsilon=7.5 \times 10^{-4}$ and $w=3$.

%Parameters $\varepsilon $ and $w$ of guided filter (in Eq.(3)) should be set carefully. For the parameter $\varepsilon$, we have empirically found that $\varepsilon \in[2.5 \times 10^{-4}, 1 \times 10^{-3}]$ generally
%yields good results  and have accordingly used $\varepsilon=7.5 \times 10^{-4}$ for the results in our experiments.
%And we fixed the parameter value of $w$ to 3 in all experiments.
% That is to say, the size of window $\omega_{k}$ is $3 \times 3$.
%Although better results could be obtained with "optimal" tuning of these parameters, from our experience with this method, it is believed that these heuristic values can achieve good results in most cases.

We compare the proposed method with four state-of-the-art algorithms: ForWaRD \cite{R.Neelamani}, TVS \cite{OlegV.Michailovich}, SV-GSM \cite{J.A.Guerrero-Colon}, L0-AbS \cite{J.Portilla} in standard test settings for deconvolution. Table.\ref{table.2} and Table.\ref{table.3} compared the results of our method and the other four methods in terms of ISNR (improvement in signal-to-noise-ratio). In our experiments, our method clearly outperforms the other four methods.
In MATLAB simulation, we have obtained times per iteration of 0.098 seconds using $256 \times 256$ image with an Pentium(R) Dual-Core CPU E5300 @2.60HZ and 2G RAM. The running time of the whole process (round 30 iterations) was 2.94s.

\begin{figure}[htbp]
  \centering  \hspace*{-1cm} \vspace*{-1cm}
  \includegraphics[width=0.5\textwidth]{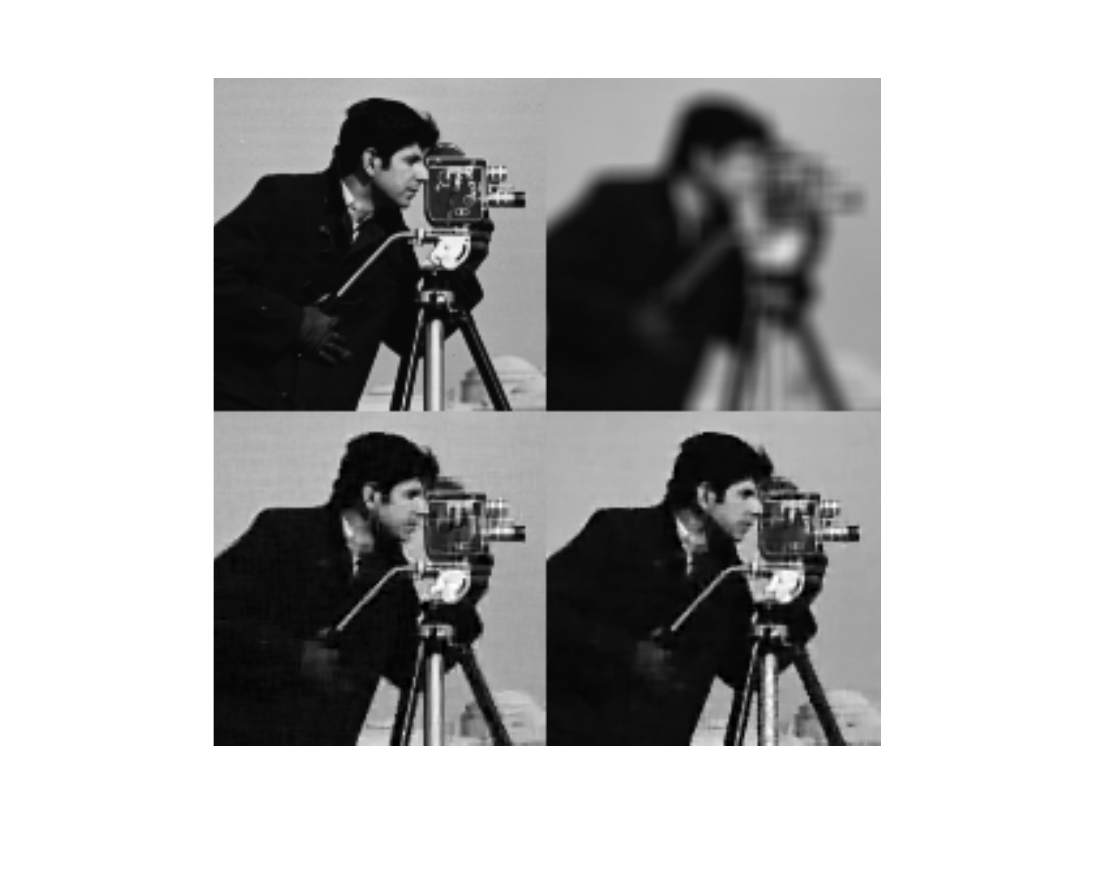}\\
  \caption{Comparisons with $Cameraman$ image in test setting 3. From left to right and from top to bottom: original image, blurred image, result from \cite{J.Portilla} (9.10 dB) and our result (9.53 dB).}\label{p.1}
\end{figure}

\begin{figure}[htb]
  \centering
  \centerline{\includegraphics[width=8.0cm]{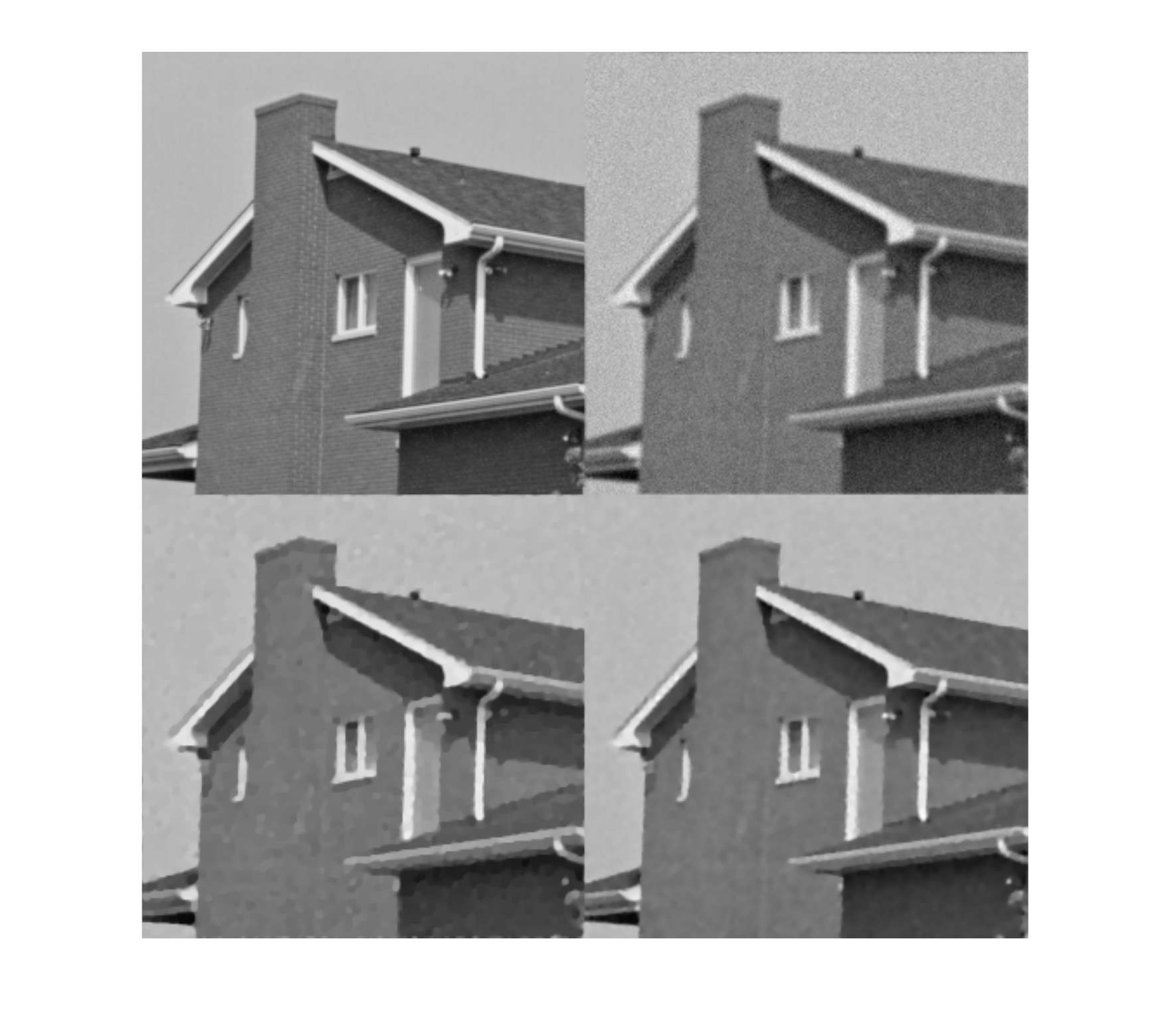}}
%  \vspace{2.0cm}
%  \centerline{(a) Result 1}\medskip
\caption{Visual comparison of $House$ image in test setting 4. From left to right and from top to bottom: original image, blurred image, result from \cite{OlegV.Michailovich} (4.49 dB) and our result (4.84 dB).}
\label{p.2}
\end{figure}

Fig.\ref{p.1} shows a visual comparison between the proposed method and the method in \cite{J.Portilla}
on $Cameraman$ image in test setting 3. Whereas both results recover the original image well, it is noticeable that our result
 preserves the edge better (see, the man's face). In Fig. \ref{p.2} we compare to \cite{OlegV.Michailovich} on $House$ image in
  test setting 4. Our method recovers the sharpness of some edges (for instance, vertical edges in window)
  that in \cite{OlegV.Michailovich} are still blurry.

\begin{table}[h]
  \begin{center}\hspace*{+0cm}
  \begin{tabular}{|c|c|c|}\hline
  Tsets & PSF & $\sigma^{2}$  \\
  \hline
  1 & $h(i,j)=1/(1+i^{2}+j^{2})$, for $i,j=-7,...,7$ &2  \\
   \hline
  2 & $h(i,j)=1/(1+i^{2}+j^{2})$, for $i,j=-7,...,7$ & 8  \\
   \hline
  3 & $h$ is a $9\times 9$ uniform kernel (boxcar) & 0.308 \\
   \hline
  4 & $h=[1\ 4\ 6\ 4\ 1]^{T}[1\ 4\ 6\ 4\ 1]/256$ & 49 \\
   \hline
  5 & $h$ is a Gaussian PSF with standard deviation 1.6 & 4 \\
  \hline
  \end{tabular}
  \end{center}
  \vspace{-0.5cm}
  {\bf \caption{Experiment settings with different blur kernels and different values of noise variance $\sigma^{2}$ for pixel values in [0,255].}\label{table.1}}
\end{table}
\vspace*{-0.5cm}
\begin{table}[h]
  \begin{center}\hspace*{0cm}
  \begin{tabular}{|c|c|c|c|c|c|c|}\hline
  Methods  & Test 1 & Test 2 & Test 3 & Test 4 &Test 5  \\
  \hline
   \cite{R.Neelamani} & 6.76  & 5.08 & 7.40 & 2.40 &  3.14  \\
    \hline
  \cite{OlegV.Michailovich} & 7.41 & 5.24 & 8.56 & 2.57 &  3.36 \\
   \hline
     \cite{J.A.Guerrero-Colon} & 7.45 & 5.55 &  7.33 & 2.73&  3.25 \\
   \hline
  \cite{J.Portilla} & 7.70 & 5.55 & 9.10 & 2.93 & 3.49 \\
   \hline
  Our Method & \textbf{8.16} &\textbf{6.09} & \textbf{ 9.53} &  \textbf{3.36} &  \textbf{3.95} \\
  \hline
  \end{tabular}
  \end{center}
  \vspace{-0.5cm}
  { \caption{ISNR (in dB) of different methods  on $Cameraman$ image.}\label{table.2}}
\end{table}

\vspace*{-0.5cm}
\begin{table}[h]
  \begin{center}\hspace*{0cm}
  \begin{tabular}{|c|c|c|c|c|c|c|}\hline
  Methods  & Test 1 & Test 2 & Test 3 & Test 4 &Test 5  \\
  \hline
   \cite{R.Neelamani} & 7.35  & 6.03 & 9.56 & 3.19 &  3.85  \\
    \hline
  \cite{OlegV.Michailovich} & 7.98 & 6.57 &  10.39 & 4.49&  4.57 \\
   \hline
     \cite{J.A.Guerrero-Colon} & 8.64 & 7.03 & 9.04 & 4.30 &  4.11 \\
   \hline
  \cite{J.Portilla} & 8.40 & 7.12 & 10.74 & 4.55 & 4.80 \\
   \hline
  Our Method & \textbf{8.83} &\textbf{7.46} &  \textbf{11.11} &  \textbf{4.84} &  \textbf{5.34} \\
  \hline
  \end{tabular}
  \end{center}
  \vspace{-0.5cm}
  { \caption{ISNR (in dB) of different methods  on $House$ image.}\label{table.3}}
\end{table}

%\begin{figure}[htb]
%
%  \centering
%  \centerline{\includegraphics[width=8.5cm]{cameraman.eps}}
%%  \vspace{2.0cm}
%%  \centerline{(a) Result 1}\medskip
%\caption{Details of $Cameraman$ image in test setting 3. From left to right an from top to bottom: original image, blurred image, image deconvolved by L0-AbS\cite{J.Portilla} (9.10 dB) and our result (9.53 dB)}
%\label{p.1}
%%
%\end{figure}

%\begin{figure}[htbp]
%  % Requires \usepackage{graphicx}
%  \centering  \hspace*{-1cm}\vspace*{-0cm}
%  \includegraphics[width=0.5\textwidth]{house.eps}\\
%  \caption{Visual comparison of $House$ image in test setting 4. From left to right an from top to bottom: original image, blurred image, image deconvolved by TVS \cite{OlegV.Michailovich} (4.49 dB) and our result (4.84 dB)}\label{p.2}
%\end{figure}

\begin{figure}[htb]

  \centering\hspace*{0.5cm}
  \centerline{\includegraphics[width=10cm]{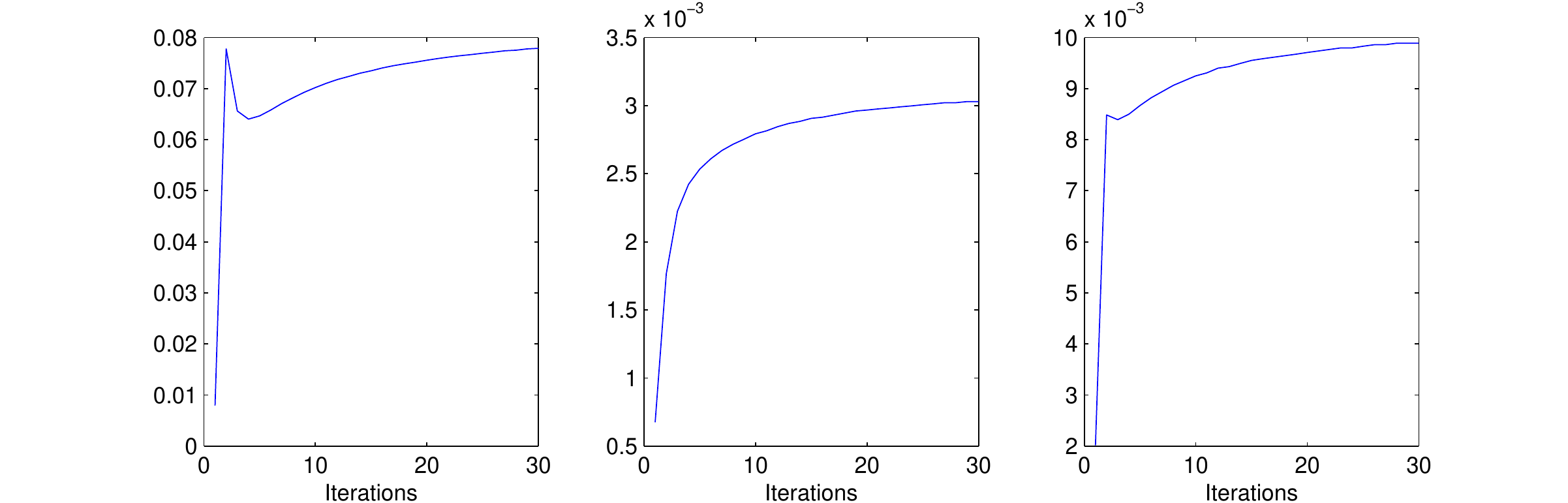}}
%  \vspace{2.0cm}
%  \centerline{(a) Result 1}\medskip
\caption{$\lambda$ values obtained by the method described in Eq.(\ref{3.8}) using $Cameraman$ image. From left to right: Test 2, Test 3, Test 5.}
\label{p.3}
\end{figure}

In Fig.\ref{p.3}, we plotted a few curves of different $\lambda$ values obtained
from Tests 2, 3 and 4 using $Cameraman$ image, respectively. Hence, unlike some of the other deconvolution
algorithms such as that in \cite{Y.Wang}, our method automatically
determines the regularization parameter at each iteration.

%\begin{figure}[htbp]
%  % Requires \usepackage{graphicx}
%  \centering  \hspace*{-1cm}\vspace*{-2cm}
%  \includegraphics[width=0.6\textwidth]{lamda.eps}\\
%  \caption{$\lambda$ values obtained by the method described in Eq.(\ref{3.8}) using $Cameraman$ image. From left to right: Test 2, Test 3, Test 5}\label{p.3}
%\end{figure}

% Below is an example of how to insert images. Delete the ``\vspace'' line,
% uncomment the preceding line ``\centerline...'' and replace ``imageX.ps''
% with a suitable PostScript file name.
% -------------------------------------------------------------------------
%\begin{figure}[htb]
%
%\begin{minipage}[b]{1.0\linewidth}
%  \centering
%  \centerline{\includegraphics[width=8.5cm]{image1}}
%%  \vspace{2.0cm}
%  \centerline{(a) Result 1}\medskip
%\end{minipage}
%%
%\begin{minipage}[b]{.48\linewidth}
%  \centering
%  \centerline{\includegraphics[width=4.0cm]{image3}}
%%  \vspace{1.5cm}
%  \centerline{(b) Results 3}\medskip
%\end{minipage}
%\hfill
%\begin{minipage}[b]{0.48\linewidth}
%  \centering
%  \centerline{\includegraphics[width=4.0cm]{image4}}
%%  \vspace{1.5cm}
%  \centerline{(c) Result 4}\medskip
%\end{minipage}
%%
%\caption{Example of placing a figure with experimental results.}
%\label{fig:res}
%%
%\end{figure}

\section{Conclusion}
We have presented a new deconvolution method based on guided image filtering. Guided filter is a novel explicit image filter.
It has been proved to be more effective than the bilateral filter in several applications. We first integrate this filter into
the deconvolution problem to propose an efficient iterative algorithm, which leads to  highquality results.
Through ten standard simulation experiments, it outperforms four existing state-of-the-art deconvolution algorithms.
We find remarkable how such a simple method with just two parameters, compares favorably to other much more sophisticated methods.
We also proposed a simple and effective method of automatically determining the regularization
parameter at each iteration.

%\vfill
%\pagebreak

%
%\section{FOOTNOTES}
%\label{sec:foot}
%
%Use footnotes sparingly (or not at all!) and place them at the bottom of the
%column on the page on which they are referenced. Use Times 9-point type,
%single-spaced. To help your readers, avoid using footnotes altogether and
%include necessary peripheral observations in the text (within parentheses, if
%you prefer, as in this sentence).
%
%
%\section{COPYRIGHT FORMS}
%\label{sec:copyright}
%
%You must include your fully completed, signed IEEE copyright release form when
%you submit your paper. We {\bf must} have this form before your paper can be
%published in the proceedings.

%\section{REFERENCES}
%\label{sec:ref}
%
%List and number all bibliographical references at the end of the paper.  The references can be numbered in alphabetic order or in order of appearance in the document.  When referring to them in the text, type the corresponding reference number in square brackets as shown at the end of this sentence \cite{C2}.
%
%% References should be produced using the bibtex program from suitable
%% BiBTeX files (here: strings, refs, manuals). The IEEEbib.bst bibliography
%% style file from IEEE produces unsorted bibliography list.
%% -------------------------------------------------------------------------
\bibliographystyle{IEEEbib}
%\bibliography{strings,refs}
%\bibliography{RN.bib}

\end{document}